\documentclass[twocolumn]{article}
\usepackage{graphicx} % Required for inserting images
\usepackage{enumitem}
\usepackage{url}
\usepackage{amsmath}
\usepackage{lipsum}
\usepackage{comment}
\usepackage{csquotes}
\usepackage{arabtex}
\usepackage{utf8}
\setcode{utf8}
\usepackage{multirow}
\usepackage{array}
\usepackage{caption} 
\usepackage {authblk}
\usepackage{hyperref}
\newcolumntype{P}[1]{>{\centering\arraybackslash}p{#1}}

\usepackage[left=0.7in,right=0.7in,top=1in,bottom=1in]{geometry}
\usepackage{cite}
\begin{document}
\title{\textbf{A multilingual training strategy for low resource Text to Speech}}
\author[1,2,*]{Asma Amalas}
\author [2,3] { Mounir Ghogho } 
\author [4] { Mohamed Chetouani } 
\author [1] { Rachid Oulad Haj Thami } 
\affil [1] {\small ENSIAS, Mohammed V University in Rabat, Morocco} 
\affil [2] {\small International University of Rabat, TICLab, Morocco}
\affil [3] {\small Faculty of Engineering, University of Leeds, United Kingdom }
\affil [4] {\small Sorbonne Universit\'e, CNRS, Institut des Systèmes Intelligents et de Robotique, Paris, France}
\affil[*]{Corresponding author: asma.amalas@uir.ac.ma}
\begin{comment}
\affil[ ]{Contributing authors: mounir.ghogho@uir.ac.ma, mohamed.chetouani@sorbonne-universite.fr, rachid.oulad@um5.ac.ma}
\end{comment}
\date{}
\maketitle
\begin{abstract}
Recent speech technologies have led to produce high quality synthesised speech due to recent advances in neural Text to Speech (TTS). However, such TTS models depend on extensive amounts of data that can be costly to produce and is hardly scalable to all existing languages, especially that seldom attention is given to low resource languages.  With techniques such as knowledge transfer, the burden of creating datasets can be alleviated. In this paper, we therefore investigate two aspects; firstly, whether data from social media can be used for a small TTS dataset construction, and secondly whether cross lingual transfer learning (TL) for a low resource language can work with this type of data. In this aspect, we specifically assess to what extent multilingual modeling can be leveraged as an alternative to training on monolingual corporas. To do so, we explore how data from foreign languages may be selected and pooled to train a TTS model for a target low resource language. Our findings show that multilingual pre-training is better than monolingual pre-training at increasing the intelligibility and naturalness of the generated speech.


\end{abstract}

\section{Introduction}
\par The need to support Text to speech (TTS) for low resource languages is not only worthwhile to business but brings also great value to social good and welfare. These systems can greatly affect the experience of certain challenged or disadvantaged groups. For example, access to written content can be challenging or even impossible for individuals with visual impairments, low literacy levels, or learning disabilities. TTS technology has the potential to empower such individuals and bridge this gap. It enables them to access and engage with written material through auditory means, thereby enhancing their human-machine interaction experiences. Currently, despite the transformative potential of TTS technology, this advantage is usually limited to speakers of well studied languages whereas TTS systems for the majority of languages and their variations are still not available and lack adequate TTS support. \newline
High-resource languages such as English \cite{ljspeech} has reached outstanding performances in TTS where generated speech became indistinguishable from human speech. However, while research for these languages has transitioned towards newer and more challenging tasks, dedicated attention to study low resource languages in general and their dialects in particular is limited.

It is a given that training deep neural networks is data hungry and require large amounts of it. This can be expensive and time-consuming especially that TTS datasets are usually high quality clean data, recorded specifically by professionals in dedicated environments with high quality equipment. These requirements exacerbate barriers faced in low resource settings. Nevertheless, the rapid and remarkable progress in deep learning motivates us to relieve the dependence of TTS systems on tens of hours of high quality paired text and speech data.
\begin{comment}
In that direction, works have either tried to construct datasets by using derivatives of already existing materials of speech corporas such as Automatic speech recognition datasets or broadcast news data, and therefore reduce the cost of building these corporas while making use of robust TTS architectures, or have tried to use deep learning techniques such as unsupervised learning \cite{unsuptts}, self-supervised learning \cite{selfsup}, semi-supervised learning \cite{semisup} \cite{semisup2} and transfer learning \cite{e2ebyCLTL}.
\end{comment}
In that direction, works have either aimed to reduce the cost of building these corporas while making use of robust TTS architectures, or have sought to use deep learning techniques. Works following the first approach have constructed datasets using derivatives of existing materials of speech corporas such as Automatic Speech Recognition (ASR) datasets or broadcast news data. Alternatively works following the second approach have employed unsupervised learning \cite{unsuptts}, self-supervised learning \cite{selfsup}, semi-supervised learning \cite{semisup} \cite{semisup2} and transfer learning \cite{e2ebyCLTL}.

Transfer learning has been a key focus area and was widely explored by scholars for low resource scenarios. For instance, in purely monolingual transfer learning, emotional TTS \cite{emotts}, speaking-style based TTS or voice cloning for a new speaker can use pre-trained models on natural speech of the same language when confronted with a limited dataset of specific speech types. Monolingual transfer learning can also sometimes include using a one speaker single high resource \cite{e2ebyCLTL} language for a different low resource language, if data in that source language is sufficient, or it includes harnessing multi-speaker data by pooling utterances of many speakers \cite{libritts}. This has been shown in many works to be more effective than speaker-dependent models trained with more data \cite{effectofmulti} even when dealing with speaker imbalanced speech corporas \cite{imbalanced}.
However, when such data isn't accessible or available, data from two or more languages and speakers can be instead leveraged, thus allowing a multilingual multi-speaker approach where the target language (low resource) can capitalize on and benefit from knowledge gained from each one of the used languages. For instance, authors of \cite{bilingual} as well as \cite{learnBiEngKo} achieved high fidelity speech when a bilingual dataset was used. 
Nevertheless, in the multilingual modeling approach, training approaches typically adopted entail co-training or multitask strategies. This usually gives better results as shown in many works namely \cite{b2s} and \cite{efficient}. Furthermore, source languages used as high resource languages are usually not carefully chosen. Although authors in \cite{acousim} address this, the approach is not tested in the multilingual setting and therefore is limited in terms of variety for similar languages. Conversely, when this is the case, that is the use of multiple close languages that supposedly share some common features, authors use languages from the same family \cite{htl} or even different families \cite{langfam} but with no prior selection.
\newline
Our aim in this work is to address the low resource scenario through a multilingual transfer learning approach where source languages are pre-selected. We investigate whether multilingual modeling for cross lingual transfer learning can be on par or better than transfer learning from one language. We compare the performance of transfer learning from the "by default" closest language to one of its dialects with the performance of leveraging multiple auxiliary close source languages from different family languages and branches. The objective is to benefit from the phonetic properties of various language families. We adopt a pipeline that merges multitask learning for the multilingual pre-training and transfer learning for the training on the single target language. Furthermore, given that our target low resource language (We consider the Moroccan Dialect as our target language: This is the informal vernacular spoken in Morocco. We refer to it in the rest of the paper by Darija) lacks any available parallel corporas for speech and text, we were compelled to explore a nontraditional data source to build a small dataset. To this end, we explore data found in the web as our speech data repository, we particularly focused on mining data from social media, specifically user-generated content from Youtube.\newline
The contributions of this work can be summarized as follows:
\begin{itemize}
\item We investigate the usability of found data on the web, in particular social media to build a TTS model for a low resource language. 
\item We explore multilingual models in cross lingual transfer learning in a low resource setting. We first present a deep learning method to accurately select source languages. We then compare the performances of the pre-training on monolingual and multilingual corporas. 
\item We explore a cascade adaptation and fine-tuning strategy to tackle the low resource setting in multilingual modeling using only 1.2 hours of parallel paired data for the target language.
\item We present the first TTS system that handles the Darija Dialect.
\end{itemize}
\section{Related work}
\subsection{Building TTS datasets using unconventional ways} 
Recently, there has been a notable shift in TTS dataset construction methodologies, driven by advancements in neural speech technologies and the pursuit of tackling more complex and challenging tasks with limited data resources. This paradigm shift has prompted researchers to explore alternative approaches to dataset creation, particularly in scenarios where traditional methods may be impractical or unfeasible. This is especially useful in the low-resource settings where datasets for certain languages are nonexistent. \newline
The LibriTTS \cite{libritts} and the CML-TTS \cite{cmltts} corporas for example are derived from original materials designed to create ASR datasets (\cite{librispeech} and \cite{mls} respectively), instead of following the traditional approach of recording. Both works followed detailed pipelines to well adapt utterances from the initial datasets. The pipelines include text and audio processing to inherit the already existing desired properties while addressing issues that could compromise TTS applications. In contrast to \cite{libritts} and \cite{cmltts} that uses audiobooks, other initiatives such as the CMU Wilderness Multilingual Speech Dataset \cite{cmu} and BibleTTS \cite{bibletts} have capitalized on publicly available Bible readings as the foundation for their speech corporas. Through meticulous filtering, cleaning, and alignment processes, these projects have demonstrated the ability to construct good synthesizers from crawled data. 
Moreover, some studies have inquired into comprehensive analyses of various types of found data to assess their suitability for TTS-style speech. For example, researchers in \cite{founddata} conducted an in-depth examination of acoustic and prosodic features to evaluate the correspondence of various types of found audio recordings with TTS requirements. Their findings suggest that audio broadcast news exhibits characteristics that closely align with the needs of TTS systems, highlighting the potential of this data source for future dataset construction efforts.
Unsupervised data selection methods were also explored as promising strategies for TTS dataset creation. In \cite{unsupdatasel}, researchers leveraged as well broadcast data and proposed an unsupervised data selection approach, achieving good results with just one hour of available data. Their approach yielded a Mean Opinion Score (MOS) of 4.4 for intelligibility, underscoring the efficacy of broadcast data coupled with selection methodologies to address the issue of dataset construction in resource-constrained scenarios.

\subsection{Low resource TTS}
Various techniques were leveraged by researchers to tackle shortage in available speech services as the majority of languages in the world lack sufficient training data. Self-supervised learning was employed to alleviate the dependence of TTS models on paired speech and text data or to enhance the models' performance by leveraging pre-trained language models such as BERT \cite{ssl1,ssl2} or by leveraging acoustic units discovered by self-supervised speech representation models such as vector-quantized variational auto-encoder (VQ-VAE) \cite{ssl3,unsup3} and HuBERT \cite{ongenspoken} to generate sequences of representations very close to phoneme sequences of speech utterances.
Furthermore, Speech chain \cite{sc} and Back transformation \cite{bt} were explored to make use of task duality between ASR and TTS, thereby mutually boosting each others performances. 
Unsupervised learning was also explored in \cite{unsuptts2} by training two modules, one for alignment and one for synthesis on non parallel datasets  to circumvent the necessity of utilizing paired data. Similarly, authors of \cite{unsuptts} leveraged unsupervised ASR models to build a TTS model.
Besides, data augmentation too can play a crucial role in mitigating the scarcity of training data in low resource TTS. Techniques like speed perturbation, pitch shifting, and adding background noise can be applied to existing speech samples, effectively increasing the dataset size without requiring additional recordings \cite{tl&da}. \newline
However, transfer learning is the standard "go-to" method to tackle various Natural Language Processing (NLP) applications for low resource languages, TTS included, whether it's used solely or combined with previous methods.  In \cite{LRSpeech} and \cite{e2ebyCLTL} the effect of TL was demonstrated using a single high resource language. However, many works such as \cite{cloning} and \cite{tl&da}, extended TL capabilities to encompass multilingual and multi-speaker modeling in the pre-training/training phase.
\newline
Further analyses were conducted on many levels to explore the effect of this multilingual modeling on the performance of synthesizers. In \cite{learnBiEngKo}, authors investigated how speech synthesis networks learn pronunciation from datasets of different languages and showed that learned phoneme embedding vectors are located closer if their pronunciations are similar across the languages. They showed that pre-training a speech synthesis model using datasets from both high and low resource languages enhances the performance of the TTS model when fine tuned on the low-resource language. In \cite{efficient}, authors investigated to what extent jointly trained multilingual multi-speaker models can be an alternative to monolingual multi-speaker models. They experimented with multiple data addition strategies for a low resource scenario and achieved improvement in terms of naturalness.
\newline
To ensure the scalability of multilingual multi-speaker models, authors of \cite{b2s} adopted a framework that can map byte inputs to spectrograms, thus addressing the mismatch of input space while dealing with multiple scripts. Their findings demonstrate the adaptability of multilingual models to new languages under extreme low-resource and few-shot scenarios. Addressing the input mismatch problem inherent in cross-lingual transfer learning, \cite{e2ebyCLTL} proposed a phonetic transformation network to learn accurate mappings between source and target languages, thereby facilitating knowledge transfer. 
\newline
On the level of language closeness, phonetic similarity was explored in \cite{htl} through a hierarchical Transfer Learning approach for low-resource languages where authors found that languages that are phonologically close can benefit significantly from transfer learning strategies in multilingual TTS models.
In \cite{acousim}, authors propose a language similarity approach that can efficiently identify acoustic cross-lingual transfer pairs across hundreds of languages. The role of family language was studied in \cite{langfam} to assess the performance of multilingual models in comparison to their monolingual counterparts. Collectively, these works highlight the significance of language similarity as a viable criteria in multilingual modeling, asserting the importance of nuanced approaches in addressing the challenges of TTS synthesis in low-resource settings.

\section{Proposed Method}
\subsection{Dataset construction}
To address the lack of available TTS corporas for Moroccan Dialect, we proceeded by building a small dataset out of audios from social media. In order to achieve effective data selection for our TTS experiments, we adhere to a set of intricate rules and steps:
\begin{itemize}[noitemsep, leftmargin=*]
    \item We carefully listen to samples of various types of data present in Youtube (Podcasts, interviews, etc) from the Moroccan content. We finally pick audios from a channel dedicated to story-telling. Although these types of videos may feature an expressive style, characterized by a wide range of prosodic variations, they are single speaker, reasonably good in voice quality and have less noise in comparison to conversational speech for example.
    \item We scrape the audios from the identified channel and exclude the ones with less than 22kHz which is a standard sample rate in TTS. 
    \item We assess the quality of speech and filter out noisy utterances by computing signal to noise ratio (SNR) using the waveform amplitude distribution analysis (WADA) \cite{wadasnr}. This method allows us to measure the strength of the speech signal to that of background noise. If WADASNR is less than 20dB, we discard the audio.
    \item We denoise the resulting set of audios to boost the quality using a source separation library.\footnotemark
    \item We split the audios into chunks lesser than or equal to ten seconds.
    \item We transcribe the audios by generating pseudo labels with the help of an ASR system.\footnotemark 
    \item We normalize text by converting numbers to their written form. We also vowelize text. Diacritics and short vowels are important in light of total absence of lexicons or phonemization tools for our target language. As the results of both transcription and vowelization have some errors, we manually fix those to ensure text accuracy.
\end{itemize}
We constructed a total of 6 hours of speech only data where each audio file is a single-channel WAV with a sample rate of 22.05kHz. A subset of 1.2 hours was used for the transcribed parallel \texttt{<audio, text>} pairs. Each utterance corresponds to a line of text in Arabic script. Since Darija has sounds not present in Standard Arabic, such as \textit{v}, \textit{p} and \textit{g}, we introduce the letters \textit{\<ڤ>}, \textit{\<پ>} and \textit{\<ڭ>} as respective corresponding graphemes in the used vocabulary to address their lack in the Arabic script. We ensured to incorporate all potential sounds found in Darija, along with their corresponding letters, into the subset of data transcriptions we used. But, it's important to note that the dataset isn't necessarily balanced. This imbalance presents an additional challenge in testing our approach and ensures that it is a representative example for extreme low-resource scenarios where data scarcity is a major concern.

Table \ref{table0} illustrates the distribution of occurrences of each letter/grapheme in the used dataset in a descending order. Given that the letters \textit{\<ڤ>} and \textit{\<پ>} are generally used in loanwords from foreign languages, we notice that they appear infrequently compared for example with long vowels that are essential in Arabic and particularly in Darija. 

\footnotetext{\url{https://github.com/deezer/spleeter}}
\footnotetext{\url{https://cloud.google.com/speech-to-text}}

\begin{table}[ht!]
\centering
\caption{Distribution of different graphemes over the transcriptions from 1.2 hours of used data}
\label{table0}
\begin{tabular}[t]{lcccccc}
\hline
token&count&token&count&token&count\\
\hline

\<ا>&8292&\<ف>&1204&\<ض>&175\\
\<ل>&5741&\<ح>&1154&\<ڭ>&160\\
\<ي>&4788&\<س>&1107&\<ث>&119\\
\<و>&3049&\<ق>&952&\<ذ>&105\\
\<ه>&2843&\<ش>&931&\<ة>&87\\
\<م>&2358&\<ج>&602&\<ء>&51\\
\<ن>&2260&\<خ>&492&\<ئ>&46\\
\<د>&2218&\<ط>&473&\<ظ>&32\\
\<ر>&1901&\<ى>&459&\<آ>&22\\
\<ب>&1820&\<ص>&394&\<إ>&16\\
\<ت>&1691&\<غ>&329&\<ؤ>&11\\
\<ع>&1385&\<أ>&235&\<پ>&5\\
\<ك>&1238&\<ز>&223&\<ڤ>&3\\
\hline
\end{tabular}
\end{table}%
\subsection{Language Selection}
In multilingual modeling, cross lingual transfer utilizes \textit{n} source languages \( \{s_1, \ldots, s_n\} \subset S \) where S is a set of languages to improve the accuracy of the downstream task performance in a low resource target language. This is an invaluable tool for many NLP applications. However, it can be unclear which languages to transfer from, and researchers tend to select these languages intuitively. One work that tackled this was proposed in \cite{acousim} where the authors proposed acoustic and non-acoustic-based features approaches to define most similar languages. However, although they scaled up their approach to handle hundreds of languages, support for many other languages and their dialectal variations remains unsolved. 
To adapt this to our use case, and instead of arbitrarily choosing these source languages for our case study, we adopted a data driven approach to compute acoustic similarity for eight languages in the same as well as across language families with regard to our target language. Unlike \cite{acousim}, we tried to simplify the model but instead act upon the input. Instead of using spectrograms or MFCCs or x-vectors for example, we use pre-trained speech embeddings extracted using Wav2Vec \cite{wav2vec}. We then train a Siamese network architecture of three branches with a ResNet backbone \cite{resnet} that takes features of Wav2Vec speech representations as input. Motivated by the fact that it leverages cross-lingual pre-training, we use the XLS-R \cite{xlsr} model that was pre-trained on 128 languages. The ResNet backbone performs 1D convolutions on input features and has 152 layers.
We use a triplet loss expressed by the following formula : \textit{L}(\( x_i \),\( x_j \),\( x_k \)) = \(max(0, d(x_i, x_j) - d(x_i, x_k) + \alpha)\) where the objective is to minimize the distance between the anchor \textit{\( x_i \)} sampled from \( \{x_1, \ldots, x_n\} \subset s_l \) and the positive example \textit{\( x_j \)} sampled from \( \{x_1, \ldots, x_n\} \subset s_l \) while maximizing the distance between the anchor and the negative example \textit{\( x_k \)} sampled from \( \{x_1, \ldots, x_n\} \subset s_m \) by a margin $\alpha$. The goal is to capture the closeness between clusters of languages by grouping utterances of a candidate closer language to our target language in the embedding space if an acoustic similarity exists. We used data from publicly available Bible recordings\footnotemark\ for the task of language classification. After training, the similarity is then calculated for the inference set on the averaged output of the last layer of the model for each language using the cosine distance.\newline
Given language family information and historical context relating to our target language, we considered eight languages for training. As we are interested in evaluating how languages of a totally different family language from that of our target language can affect transfer learning, the majority of the languages used for training are from another family language with diverse subdivisions.
Figure 1 provides an overview of the model used, while Table \ref{tab:my-table-2} details the training details.

\begin{table}[]
\caption{Languages used for selection and their family groups. \textit{Utt.} is the number of utterances used for each language during training and inference.}
\label{tab:my-table-2}
\begin{tabular}{|c|c|c|P{1.2cm}|}
\hline
Family & Subdivision              & Language   & \textit{Utt.} \\ \hline
\multirow{6}{*}{Indo-European} & \multirow{2}{*}{Germanic} & English & \multirow{8}{*}{1650} \\ \cline{3-3}
       &                          & Dutch      &            \\ \cline{2-3}
       & \multirow{3}{*}{Romance} & French     &            \\ \cline{3-3}
       &                          & Spanish    &            \\ \cline{3-3}
       &                          & Portuguese &            \\ \cline{2-3}
       & Iranian                  & Farsi      &            \\ \cline{1-3}
\multirow{2}{*}{Afro-Asiatic}  & \multirow{2}{*}{Semitic}  & Hebrew  &                       \\ \cline{3-3}
       &                          & Amharic    &            \\ \hline
\end{tabular}
\end{table}
\footnotetext{https://www.faithcomesbyhearing.com}
\begin{figure}[!h] 
    \centering
    \includegraphics[scale=0.5]{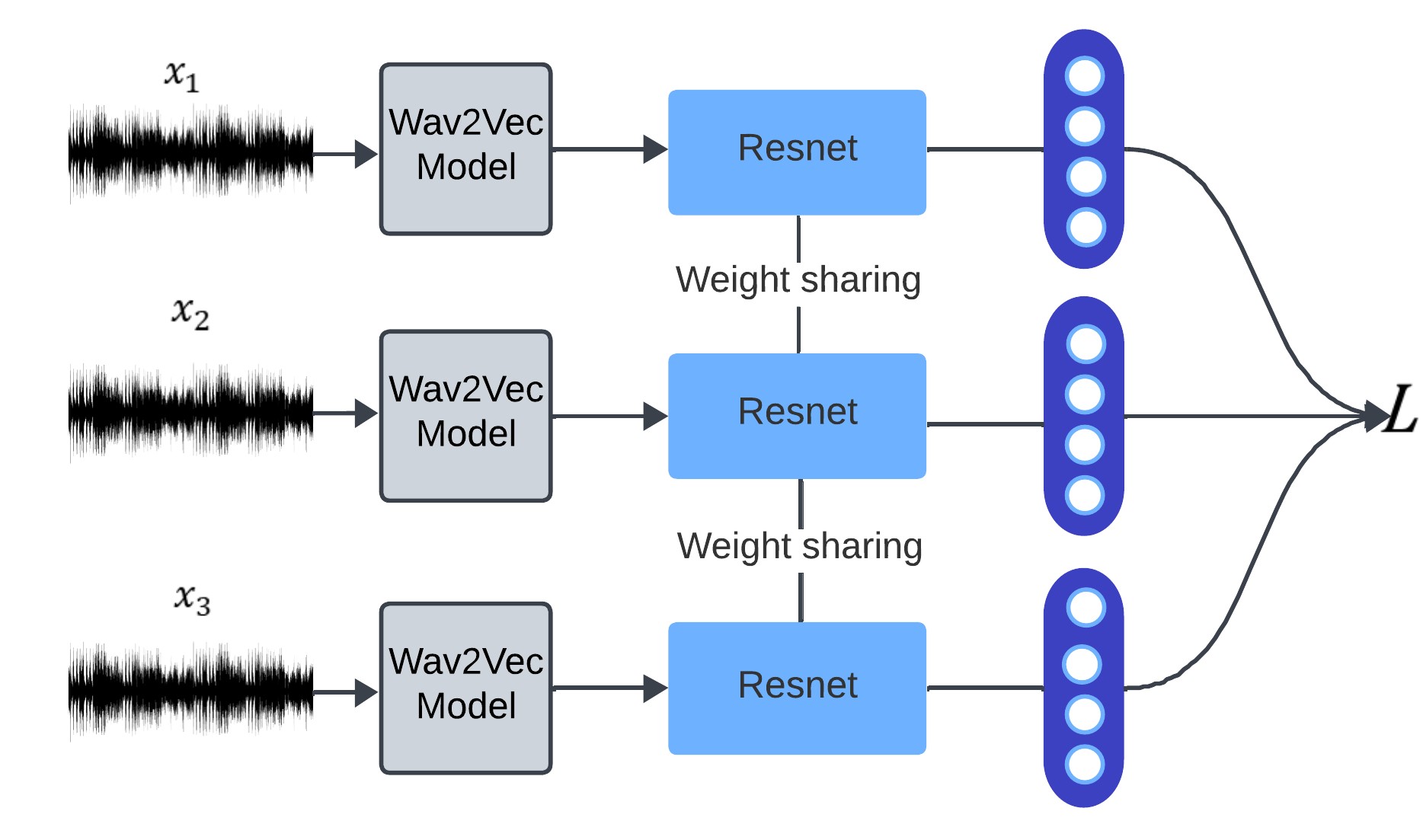}
    \caption{Overview of the adopted architecture for the language similarity model.}
    \label{fig:langsimoverview}
\end{figure}
\subsection{Adopted TTS Architectures}
Modern TTS systems consist of three basic components: a text analysis module that converts a text to linguistic features, an acoustic model that generates acoustic features from linguistic features, and then a vocoder that synthesizes a waveform from acoustic features. The choice of the adequate architecture depends on the working mechanisms of these three elements especially when dealing with small amounts of data that imposes many constraints.\newline
State of the art of neural models designed for TTS has reached great performances and offers a wide range of possibilities to pick from. End-to-end architectures have gained popularity since their inception for various sequence to sequence tasks including TTS. In that regard, models such as Tacotron2 \cite{tacotron}, TransformerTTS \cite{transformertts} and FastSpeech2 \cite{fastspeech2} were presented and explored by researchers. \newline
Tacotron2 leverages an encoder-attention-decoder framework to output mel-spectrograms from character inputs. It is based on a convolutionnal neural network and a bidirectionnal Long Short-Term Memory (LSTM) layer encoder and a recurrent neural network (RNN) of LSTM units decoder. The encoder and decoder are bridged by an attention network that employs location-sensitive attention \cite{attlocsen}. Additionally, Tacotron uses a 5-layer convolutional postnet to refine the synthesized speech.\newline
In order to mitigate issues related to efficiency in training and inference from one hand, and to modeling long dependencies in text and speech sequences that RNNs (e.g Tacotron2) suffer from, authors of \cite{transformertts} leverage the transformers architecture and self-attention to generate mel-spectrograms from phonemes. The paper presents similar voice quality with Tacotron2 while resolving the problem of slow training. However, the model's robustness can be compromised due to parallel computation which speeds training, and therefore, the generated speech can suffer from word skipping and word repeating which mainly results from inaccurate attention alignments between text and mel-spectrograms in encoder-attention-decoder based autoregressive generation. To address this,  some works such as \cite{fastspeech} propose instead non-autoregressive models to tackle said robustness issues. The authors propose to remove the attention mechanism between text and speech and use instead a length regulator to bridge the length mismatch between text and speech by leveraging a duration predictor to predict the duration of each phoneme. The model works under the teacher-student training pipeline to enhance its controllability and robustness. This is extended later on in \cite{fastspeech2} by introducing more variation information of speech (e.g., pitch, energy and more accurate duration) as conditional inputs. Due to its non-autoregressive nature, FastSpeech2 attends to the exposure bias and error propagation problems in autoregressive generation. In addition, given that it adopts a feed-forward Transformer network to generate mel-spectrograms in parallel, it significantly reduces inference time. 
\newline
The present study deals with a low resource scenario, and therefore some conditions had to be considered in order to choose the right model to test our approach. Firstly, the model needs to simplify the text analysis module to lessen requirement on human preprocessing and linguistic feature development as this is challenging for a low resource language. Secondly, the model needs to learn the alignments between text and speech sequences in an unsupervised manner as this offers more flexibility and adaptability. Thirdly, the model needs to be end-to-end to reduce reliance on feature engineering, to simplify the training process and to avoid error propagation commonly encountered in models that are not end-to-end.
In light of all of the above, we conducted our experiments using mainly TransformersTTS and FastSpeech2 as they're conform with our use case. These models can work with raw text (characters) as input, can learn alignments between speech and text and adopt end-to-end architectures.

\begin{figure*}[t]
    \centering
    \includegraphics[scale=0.66
    ]{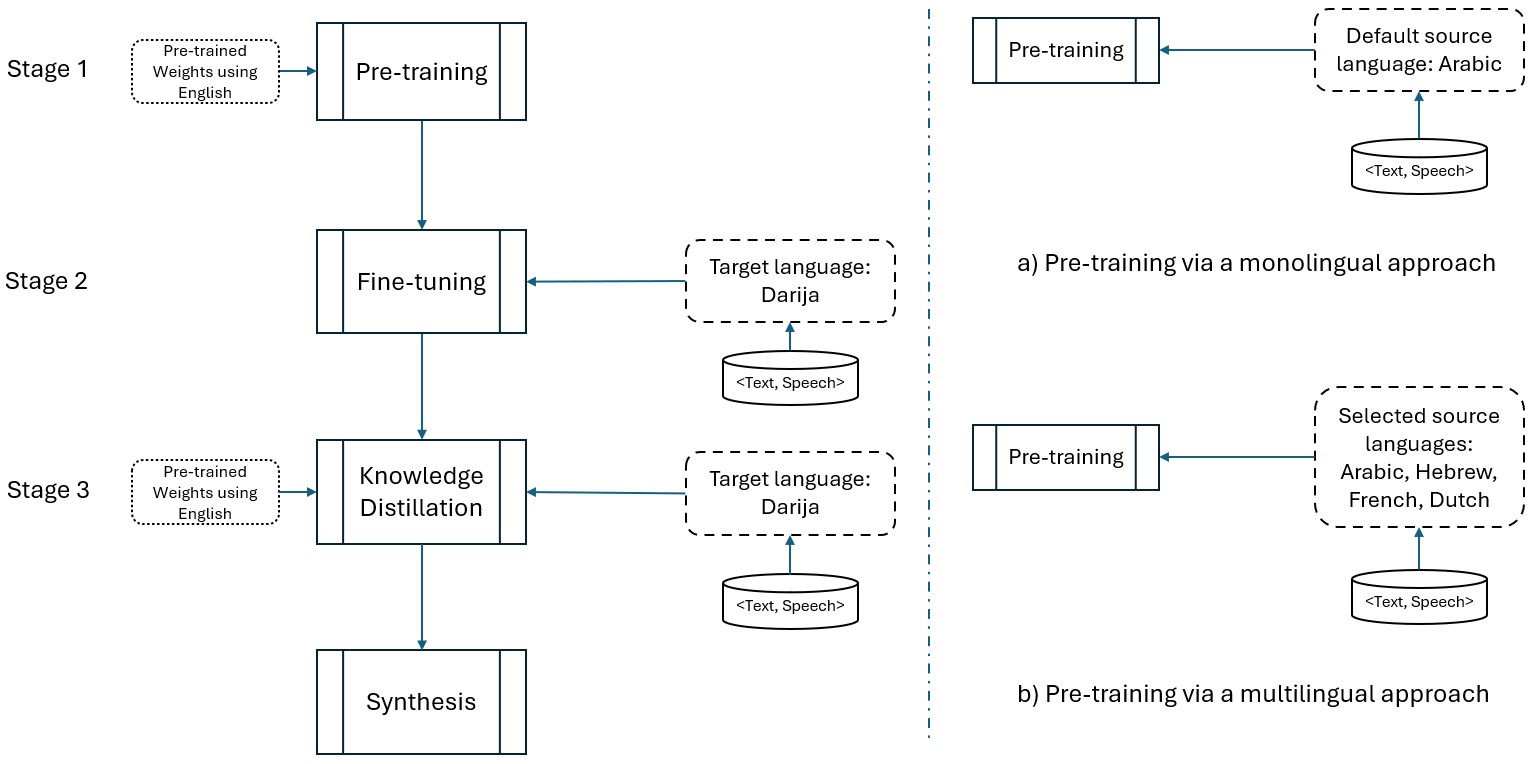}
    \caption{Three stage adaptation pipeline on target language. (a) We perform pre-training using the default closest language for the monolingual approach. (b) We perform pre-training using multiple languages selected via the model in Figure \ref{fig:langsimoverview} for the multilingual approach}
    \label{fig:finetunepipeline}
\end{figure*}
\section{Experiments and Results}
\subsection{Experimental design and training procedure}

This study aims to investigate the effect of monolingual and multilingual corporas on the performance of a fine-tuned model for a low resource target language using data found on the web, especially social media. We design two main experiments that we use to compare pre-training on one language and pre-training on several languages. We condition these experiments on the amount of used data, meaning that if \textit{h} hours of the selected single source language was used for the first case, then the same amount, divided equally on the selected source languages for the second case is used when multiple source languages are employed. Using our language selection method, we identify the top three most similar languages by computing the cosine distance between language pairs. We consider pairs (\textit{targetLang},\textit{\( s_i \)}) where \textit{\( s_i \)} belongs to \textit{S}. We try to include one language from each family subdivision for diversity. For our source languages, we end up selecting two languages from the same language family (Afro-Asiatic) as our target language, one is set by default, that is Arabic, and the other is Hebrew. Two other languages are from a different language family (Indo-European) with two distinct subdivisions, French as romance and Dutch as Germanic.  
\newline
The adaptation pipeline used is illustrated in Figure \ref{fig:finetunepipeline}. We follow a sequential fine tuning strategy that consists of a pre-training stage, a fine-tuning stage, and a knowledge distillation stage. The motivation behind the pre-training stage is knowledge transfer from multilingual multi-speaker data. This stage is preceded with a weights' initialization step where the motivation is parameter generalization including alignment maps. The fine-tuning stage concerns adapting the model on the downstream task while the knowledge distillation stage is dedicated to fix errors. The two first stages are carried out using the TransformerTTS model while the third stage uses FastSpeech2 model. In this stage, the model is refine-tuned using extracted ground-truth duration of each input token (character in our case) from the attention weights of the autoregressive teacher model with teacher forced prediction.
\newline
Our first experiment was designed to investigate the performance of the fine-tuning pipeline on the target language using a single high resource language. For this purpose, we train both autoregressive models using 12 hours of Arabic data (referred to as \(\text{MONO}_{\text{Ar}}\)) and then fine tune on one hour and a quarter of Darija . We use Arabic as our source language given that Darija is considered one of its dialects and therefore Arabic is chosen by default for cross lingual transfer learning in this experiment. Afterwards, we refine-tune using knowledge distillation on the same Darija dataset (referred to as \(\text{MONO}_{\text{Dar}_\text{1}} \)).
\newline
Our second experiment was designed to investigate the performance of the pipeline on the target language. We use the languages previously chosen using our language selection method as our source languages. In addition to these, we also add Arabic given that it is the default closest language. We train the autoregressive model (referred to as \( \text{MULTI}_{\text{Ar}, \text{Heb}, \text{Fr}, \text{Du}}\)) using a concatenation of data from the four languages. A total of 12 hours of Multilingual multi-speaker data is used where each language contributes to the dataset by a portion of three hours. As we aren't interested in language or speaker information, we don't use language ID vectors or speaker embeddings in training, the languages are thus modeled using a single encoder and using the datasets simultaneously while training. The resulting model is then fine-tuned following stages 2 and 3 of the pipeline on one hour and a quarter of Darija (referred to as \(\text{MONO}_{\text{Dar}_\text{2}} \)).
Given that 3 hours per language was insufficient for training from scratch in the multilingual modeling, we first initialize with weights from a pre-trained English model (referred to as \(\text{MONO}_{\text{En}}\))  for all adopted architectures. To ensure a fair comparison between \(\text{MONO}_{\text{Dar}_\text{1}} \) and \(\text{MONO}_{\text{Dar}_\text{2}} \), we also initialize the models in the first experiment. We opted for English given that it is the widely studied language and has high quality resources.\newline
All models were trained using character inputs. The sampling rate was set to 22.05kHz. An 80-dimension of Mel filter bank, 1024 samples of FFT length, and 256 samples of frame shift were used for speech analysis. We used a Transformer encoder and a Transformer decoder, both consisting of 6 blocks. The postnet has five convolutional layers with a kernel size of five. The dimension of the attention was set to 512 and the number of attention heads to 8. we trained the model for 200 epochs in stages 1 and 2 and 400 epochs in stage 3. We used the Noam optimizer with the learning rate and warm-up step set to 1.0 and 8000, respectively. To improve the training efficiency, we used guided attention loss \cite{guidedloss}.\newline
We used LJSpeech \cite{ljspeech} for \(\text{MONO}_{\text{En}}\), ClarTTS \cite{clartts} for \(\text{MONO}_{\text{Ar}}\),  a combination of subsets from ClarTTS, SASpeech \cite{hebrew} and CSS10 \cite{css10} for \( \text{MULTI}_{\text{Ar}, \text{Heb}, \text{Fr}, \text{Du}}\). The details of used subsets are illustrated in Table \ref{tab:my-table}.\newline
The models \(\text{MONO}_{\text{Dar}_\text{1}} \) and \(\text{MONO}_{\text{Dar}_\text{2}} \) in stages 2 and 3 were fine-tuned after excluding the parameters of the embedding layer due to token discrepancy resulting from different language scripts.\newline
For the vocoder module, we experimented with the GriffinLim algorithm \cite{griffinlim} as well as Parallel WaveGAN (PWG) \cite{pwg} to generate waveforms from the output of text to mel-spectrogram networks. The PWG model is based on a generative adversarial network. We opt for PWG as it can be easily trained due to its compact architecture all while retaining the capability to produce high-fidelity speech. We train the model using 6 hours from our constructed dataset for 600k steps. We also used a pre-trained PWG model for synthesis (referred to as \(\text{PWG}_{\text{pre-trained}}\)) as we found that it gives better results. All our Experiments were implemented using the ESPnet toolkit\footnotemark. We used the public implementation\footnotemark\ to train the PWG neural vocoder.
\footnotetext{https://github.com/espnet/espnet}
\footnotetext{https://github.com/kan-bayashi/ParallelWaveGAN}
%p{"column width"} %align top vertically
%m{"column width"} %align middle vertically
%b{"column width"} %align bottom vertically
% \usepackage{multirow}
\renewcommand{\arraystretch}{1.4}
\begin{table*}[h]
\centering
\caption{Duration (\textit{in minutes}) and number of utterances used in each stage and each language for training, development and test sets. \textit{Init.} refers to the step of weights' initialization }

\label{tab:my-table}
\begin{tabular}{|P{1.4cm}|P{2.2cm}|P{1.4cm}|P{1.4cm}|P{1.4cm}|P{1.4cm}|P{1.4cm}|P{1.4cm}|}
\hline
\multicolumn{1}{|c|}{\centering Stage} & \multicolumn{1}{c|}{\centering Dataset} & \multicolumn{1}{c|}{\centering Language} & \multicolumn{1}{c|}{\centering Duration} & \multicolumn{1}{c|}{\centering Speaker} & \multicolumn{1}{c|}{\centering Train} & \multicolumn{1}{c|}{\centering Dev.} & \multicolumn{1}{c|}{\centering Test} \\ \hline
\textit{Init.}     & LJSpeech\cite{ljspeech}              & English & 1440 & Female & \multicolumn{1}{c|}{12600} & 250                 & 250                 \\ \hline
\multirow{4}{*}{1} & ClarTTS\cite{clartts}                 & Arabic  & 288.69 & Male   & 4029                       & \multirow{4}{*}{100} & \multirow{4}{*}{100} \\ \cline{2-6}
                   & SASpeech\cite{hebrew}                 & Hebrew  & 277.64 & Male   & 2948                       &                     &                     \\ \cline{2-6}
                   & \multirow{2}{*}{CSS10\cite{css10}} & Dutch   & 288.71 & Male   & 2046                       &                     &                     \\ \cline{3-6}
                   &                        & French  & 241.12 & Male   & 1844                       &                     &                     \\ \hline
2+3                & In-house               & Darija  & 76.36 & Female & 464                        & 25                  & 25                  \\ \hline
\end{tabular}
\end{table*}
\subsection{Results}
We experimented with both Tacotron2 and TransformersTTS. However, we only report results related to models trained using TransformersTTS as we found that Tacotron2 didn't yield good audio quality for stages 2 and 3. In order to evaluate the performance of our models and assess the quality of the synthesized speech, we use the objective evaluation of the Mel Cepstral Distortion (MCD) as metric in all of our experiments. Models are evaluated on the test set containing 2.8 minutes of audio from 25 utterances.\newline
We conduct an ablation study as depicted in Table \ref{table1} and Table \ref{table2}  to evaluate the impact of each stage of the pipeline on the resulting speech. We consider the baseline to be the model trained up to stage 2 with no vowelization and using the GriffinLim algorithm for synthesis. 
\newline
\newline
\textbf{Exp1 : \( \text{MONO}_{\text{En}} \rightarrow \text{MONO}_{\text{Ar}} \rightarrow \text{MONO}_{\text{Dar}_\text{1}} \)}
\newline
\par
The first experiment uses a single language in stage 2. The results in Table \ref{table1} show that the MCD decreases with each stage of forward finetuning of the pipeline. knowledge distillation (KD) in particular affects the results mostly. The addition of diacritics further improves results. We also notice that the neural vocoder (PWG) significantly boosts the quality of synthesis compared to the GriffinLim algorithm.
\begin{table*}[ht!]
\centering
\caption{Objective evaluation for experiment 1}
\label{table1}
\begin{tabular}[t]{lcc}
\hline
Method&MCD[db]\\
\hline
Baseline+GL&12.85 ± 1.30\\
Baseline+Diacritics+\(\text{PWG}_{\text{pre-trained}}\)&9.12 ± 0.84\\
\begin{comment} 
Baseline+KD+GL&13.6861 ± 1.2222\\ 
\end{comment}
Baseline+Diacritics+KD+\(\text{PWG}_{\text{pre-trained}}\)&8.93 ± 1.12\\
\hline
\end{tabular}
\end{table*}%
\newline
\newline
\newline
\newline
\textbf{Exp2 : \( \text{MONO}_{\text{En}} \rightarrow \text{MULTI}_{\text{Ar}, \text{Heb}, \text{Fr}, \text{Du}} \rightarrow \text{MONO}_{\text{Dar}_\text{2}} \)}
\newline
\begin{table*}[ht]
\centering
\caption{Objective evaluation for Experiment 2}
\label{table2} 
\begin{tabular}[t]{lcc}
\hline
Method&MCD[db]\\
\hline
Baseline+GL&12.82 ± 1.29\\
Baseline+Diacritics+\(\text{PWG}_{\text{pre-trained}}\)&8.98 ± 0.78\\
\begin{comment}
Baseline+KD+GL&13.2724 ± 1.2148\\
\end{comment}
Baseline+Diacritics+KD+\(\text{PWG}_{\text{pre-trained}}\)&8.64 ± 0.89\\
\hline
\end{tabular}
\end{table*}%

\begin{table*}[ht!]
\centering
\caption{Objective evaluation for custom PWG model in Experiments 1 and 2}
\label{table3}
\begin{tabular}[t]{lccc}
\hline
Method&MCD[db]&DNSMOS\\
\hline
\(\text{Baseline}_{\text{\textit{Exp1}}}\)+Diacritics+KD+\(\text{PWG}_{\text{custom}}\)&9.15 ± 1.03&3.19 ± 0.27\\
\(\text{Baseline}_{\text{\textit{Exp2}}}\)+Diacritics+KD+\(\text{PWG}_{\text{custom}}\)&9.03 ± 0.94&3.30 ± 0.28 \\
\hline
\end{tabular}
\end{table*}%
The same pattern is noticed in terms of results for the second experiment when multiple languages are used. We can notice that continuing fine-tuning up to stage 3  enhances the quality of the generated speech. However, results from \textit{Exp 2} are slightly better than \textit{Exp 1} in all of the stages especially when diacritics are added to the text in the training dataset. These results match the findings in \cite{unsupdatasel} that also found that vowelization helps to enhance the performance of an Arabic TTS model.
\newline
\newline
\indent In order to evaluate the effect of the non-autoregressive model trained in stage 3 for both experiments, we calculated the number of substitutions, insertions and deletions in the test set. Usually, this is done by comparing the ground truth text with the transcriptions from an ASR system, but given that available systems for our target language are not entirely accurate, we decided to evaluate the results manually by listening to utterances from the test set. Table \ref{table4} shows that knowledge distillation in the non autoregressive generation reduces tremendously the number of substituted, inserted and deleted words in the generated speech, both in monolingual and multilingual settings, with the model \(\text{MONO}_{\text{Dar}_\text{2}} \)  outperforming the model \(\text{MONO}_{\text{Dar}_\text{1}} \) especially in terms of deletions. These results show the potential of models such as FastSpeech2 in adjusting errors and fixing pronunciation from autoregressive modeling when small amounts of data are used. \newline
All results from the experiments show that adding data from foreign auxiliary languages has a positive effect on the quality of the generated speech. When we compare the models trained on monolingual and multilingual corporas in stages 2 and 3, either in terms of naturalness or intelligibility, we notice that the diversity of multilingual corporas contributes in bettering the model's robustness and therefore improves the results of generated speech. The model benefits from language variety and speaker variety, and thus learns better prosody and pronunciation.\newline
Furthermore, these findings show as well the potential of the data reservoir present in social media as an option to create TTS datasets instead of conventional ways. Generated speech in inference has a reasonable quality and can be further pruned to include more variability in terms of domains, speaking styles and accents to ensure wider adaptability and better naturalness. The usability of this data is also demonstrated for the synthesis part. Using this type of data, we can successfully train neural vocoders, namely the PWG model. Table \ref{table3} shows the objective evaluations of a custom vocoder \(\text{PWG}_{\text{custom}}\). Although the pre-trained version on English gave slightly better results, considering that it was trained on a bigger dataset and longer than our custom model (24 hours vs 6 hours and 3M iterations vs 600k iterations), this still shows the potential of using data found on social media to train such synthesis models especially that the generated speech from the custom PWG remains natural and intelligible as shown by MOS values. 
\begin{table}[ht]
\centering
\caption{Comparison of number of substitutions, insertions and deletions in models before and after Knowledge Distillation}
\label{table4} 
\begin{tabular}[t]{lccc}
\hline
Model&Sub.&Ins.&Del.\\
\hline

\(\text{MONO}_{\text{Dar}_\text{1}}\)&14&27&23\\
\(\text{MONO}_{\text{Dar}_\text{1}}\)(with KD)&2&1&14\\

\(\text{MONO}_{\text{Dar}_\text{2}}\)&8&34&27\\
\(\text{MONO}_{\text{Dar}_\text{2}}\)(with KD)&4&1&2\\
\hline
\end{tabular}
\end{table}%

Subjective evaluations such as mean opinion score (MOS) are a de-facto standard for speech synthesis and are commonly used in TTS to assess an audio's quality. To evaluate the perceptual quality of the synthesized speech, the MOS metric is measured using feedback from human annotators. We conducted a subjective listening test using the MOS methodology. We compared the two models \(\text{MONO}_{\text{Dar}_\text{2}}\) (with KD) and \(\text{MONO}_{\text{Dar}_\text{1}}\) (with KD) using 24 sentences from the test set. A panel of ten native speakers participated in the evaluation where each subject evaluated twelve samples of each model (24 samples in total) and rated the intelligibility and naturalness of each sample on a 5-point scale: 5 for excellent, 4 for good, 3 for fair, 2 for poor and 1 for bad.

\begin{table}[ht]
\centering
\caption{Subjective evaluation results using Mean Opinion Score}
\label{table8} 
\begin{tabular}[t]{lccc}
\hline
Model&Intelligibility&Naturalness\\
\hline
\(\text{MONO}_{\text{Dar}_\text{1}}\)(with KD)&2.86 ± 0.45&2.95 ± 0.42\\
\(\text{MONO}_{\text{Dar}_\text{2}}\)(with KD)&3.72 ± 0.47 &3.44 ± 0.47 \\
\hline
\end{tabular}
\end{table}%
Given that human assessment is time consuming we considered an automatic approach for the subjective evaluation of the custom Parallel WaveGAN \(\text{PWG}_{\text{custom}}\). Various studies attempted to address this by building deep learning models that can predict MOS scores. We found that \cite{dnsmos} predicts acceptable MOS values. We therefore primarily resort to this measure as our automatic subjective metric. We rely on the DNSMOS model for the MOS values as we found that it generalizes well and better than the other models for out-of-domain (OOD) data, other works such as \cite{unsupdatasel} have also found that it achieved high correlation with human annotators. The work in \cite{dnsmos} was part of the deep noise suppression (DNS) Challenge series where the authors developed a model for automatic quality assessment. The model takes into account speech quality as well as background noise quality to predict the overall quality of an audio mimicking thus the subjective metric MOS obtained from human listening tests.
\newline
As shown in Table \ref{table8}, \(\text{MONO}_{\text{Dar}_\text{2}}\) achieves a MOS of 3.72 in intelligibility and a MOS of 3.44 in naturalness, which outperforms the \(\text{MONO}_{\text{Dar}_\text{1}}\) system. This demonstrates the significant improvement brought by multilingual modeling for both criteria. Given the amount and type of data used for the target language, these results are promising and suggest potential for future enhancements.

\section{Conclusion}
This work aimed to investigate the effectiveness of multilingual modeling in improving TTS systems for low-resource language neural speech synthesis. Our results showed that the addition of auxiliary non-target language data from different language families can positively impact the quality of generated speech in a low resource language. In a scenario where the target language is a dialect of another language, multilingual modeling can even be a viable alternative to its monolingual counterpart when data in that language is not readily available. The models trained on multilingual corporas show better generalization capabilities compared to models trained on monolingual corporas even when fine tuning is done with data not specifically tailored for TTS. This conclusion highlights the robustness and versatility of multilingual modeling approaches in low-resource settings. By leveraging diverse linguistic resources and embracing a multilingual approach, more effective, holistic and inclusive speech synthesis technologies can be developed to suit the needs of diverse language communities.
\newline
\textbf{Limitations and future work}: The proposed approach is not without limitations. The language selection model could be extended and enhanced by scaling up the training data in order to include more languages. In this way, the model is more likely to offer a broader phonemic variety for source languages. Also, the data available for our default source language has constrained the amount we used in pre-training for both approaches, leading us to use pre-trained weights from an English model. Ideally, this step should have been unnecessary.  \newline
For future works, we intend to focus on experimenting with imbalanced datasets for source languages and investigate how that would affect the overall performance. We could also extend the selection model to work at the utterance level instead of the language level.
\newline

{\bibliography{references}}
\small\bibliographystyle{vancouver}
\end{document}